\documentclass[letterpaper, 10 pt, conference]{ieeeconf}  

\usepackage{cite}
\usepackage{amsmath,amssymb,amsfonts}
\usepackage{algorithm}
\usepackage{algorithmic}
\usepackage{graphicx}
\usepackage{textcomp}

\usepackage{booktabs}
\usepackage{multirow}
\usepackage{tabularray}
\usepackage{comment}
\usepackage{float} 
\usepackage{booktabs}
\usepackage{makecell}
\usepackage{pifont}

\usepackage[table]{xcolor}
\usepackage{adjustbox}
\usepackage{amsmath,amssymb,booktabs,tabularx,array}
\usepackage{xurl}
\usepackage{hyperref}
\usepackage{xurl}
\hypersetup{hidelinks}
\IEEEoverridecommandlockouts                              

\newcommand{\rewardcell}[3]{%
  \begin{minipage}[t]{\linewidth}
    \vspace{0pt}
    \parbox[t]{0.67\linewidth}{%
      \raggedright\bfseries #1%
    }%
    \hfill%
    \parbox[t]{0.30\linewidth}{%
      \raggedleft
      \adjustbox{max width=\linewidth}{#2}%
    }%
    \par
    \vspace{0.25ex}
    \noindent
    \adjustbox{max width=\linewidth}{%
      $\displaystyle #3$%
    }%
    \par
  \end{minipage}%
}

\title{\LARGE \bf
Thor: Towards Human-Inspired Whole-Body Reactions for Intense Contact-Rich Environments
}

\author{%
  \parbox{0.97\textwidth}{%
    \centering
    \normalsize
    Gangyang Li$^{1,2}$, Hongzhe Shi$^{1,3}$,
    Qing Shi$^{2}$, Youhao Hu$^{1}$,
    Cong Ma$^{4}$, \\
    Zhongyuan Wang$^{1}$,
    Xinlong Wang$^{1}$ and Shaqi Luo$^{1,*}$
    \\[0.6em]
    \small
    $^{1}$Beijing Academy of Artificial Intelligence
    $^{2}$Beijing Institute of Technology
    \\[0.2em]
    $^{3}$Tsinghua University
    $^{4}$XYZ Embodied AI
  }%
}

\begin{document}

\makeatletter
\g@addto@macro\@maketitle{%
  \par
  \noindent
  \begin{minipage}{\textwidth}
    \centering
    \includegraphics[width=1.0\linewidth]{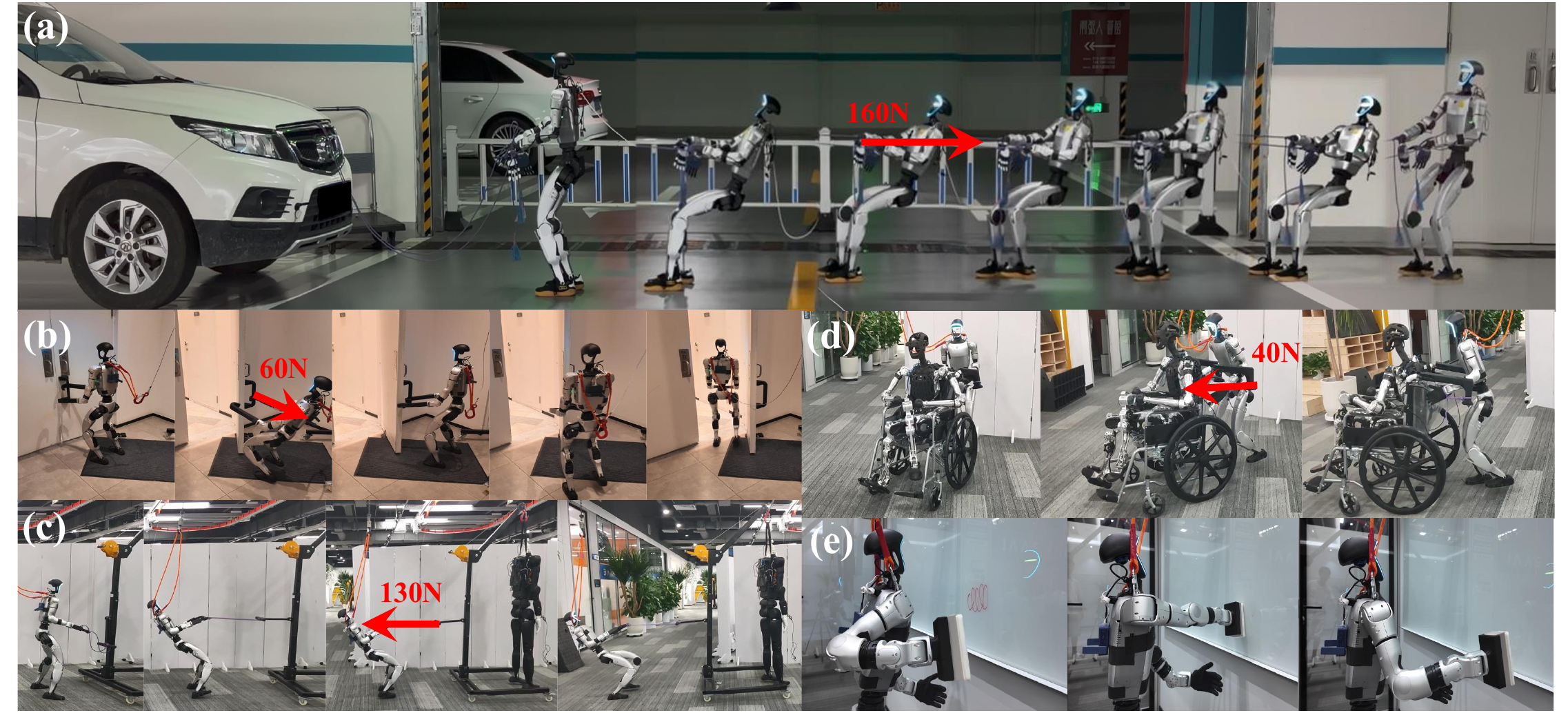}

    \def\@captype{figure}
    \caption{Humanoids performing tasks involving forceful
    interactions with the environment:
    (a) towing a 1.7-ton passenger car;
    (b) opening a fire door with one hand, requiring
    approximately 60 $N$ of pulling force;
    (c) pulling a rack carrying a 70 $kg$ load, 
    requiring approximately 130 $N$ of pulling force;
    (d) pushing a wheelchair carrying a 60 $kg$ robot to make a turn;
    (e) wiping a whiteboard with one hand.}
    \label{fig1}
  \end{minipage}
  \par
  \vspace{0.6em} 
}
\makeatother

\maketitle
\setcounter{figure}{1}
\thispagestyle{empty}
\pagestyle{empty}

\begin{abstract}
Maintaining whole-body stability and motion tracking under
large interaction forces remains challenging for humanoids.
We present Thor, a reinforcement learning framework for
forceful humanoid loco-manipulation.
Thor jointly trains lower-body, waist, and upper-body policies
with shared whole-body observations and body-specific rewards
to coordinate locomotion and force adaptation, waist posture
regulation, and upper-body motion tracking.
We further introduce a force-adaptive torso-tilt (FAT2)
objective that derives a load-dependent horizontal
center-of-mass offset reference from quasi-static moment
balance.
Capacity-matched simulation ablations show that the
three-policy architecture improves tracking under large
external force disturbances, while real-world ablations
demonstrate that FAT2 increases peak pulling capability.
On the Unitree G1, Thor achieves mean peak dual-hand
pulling forces of $(167.7 \pm 2.4)\,\mathrm{N}$ and
$(145.5 \pm 2.0)\,\mathrm{N}$ during backward and forward
locomotion, exceeding the best-performing baseline by
$68.9\%$ and $74.7\%$, respectively.
Real-world demonstrations include opening a fire door
with one hand using approximately $60\,\mathrm{N}$ of
pulling force and towing a $1.7$-ton passenger car.
\end{abstract}

\section{Introduction}

Humanoids show promise in service \cite{Hold_My_Beer}, industrial \cite{Opt2Skill_Imitating}, and disaster-response applications \cite{2016Humanoids_Motion_generation}, where forceful interactions require whole-body stability \cite{2024TRO_Adaptive_Force_Based}.
For example, opening a fire door with one hand requires sustained pulling and coordinated backward stepping while counteracting the resulting torsional moments.
Model-based methods \cite{2024TRO_Adaptive_Force_Based,2021RAL_Humanoid_Loco-Manipulations,2016RAL-Interaction_Force_Reconstruction,2018TRO-Quadratic_Programming,2019RAL-Torque-Based_Balancing} often rely on accurate dynamics models and measured or estimated external forces, which can limit adaptability to unknown interactions.
Reinforcement learning (RL) \cite{Falcon,2024ICRA-Sim-to-Real,2024ICRA-Learning_Force_Contro,2024CoRL-Human_Plus,2025ICRA-Mobile-TeleVision} offers adaptive policies learned through environmental interaction. However, maintaining stability and accurate motion tracking under large external forces remains challenging, requiring coordinated locomotion, postural adaptation, and upper-body task execution.

\begin{figure*}[!t]
\centering
\includegraphics[width=0.8\textwidth,]{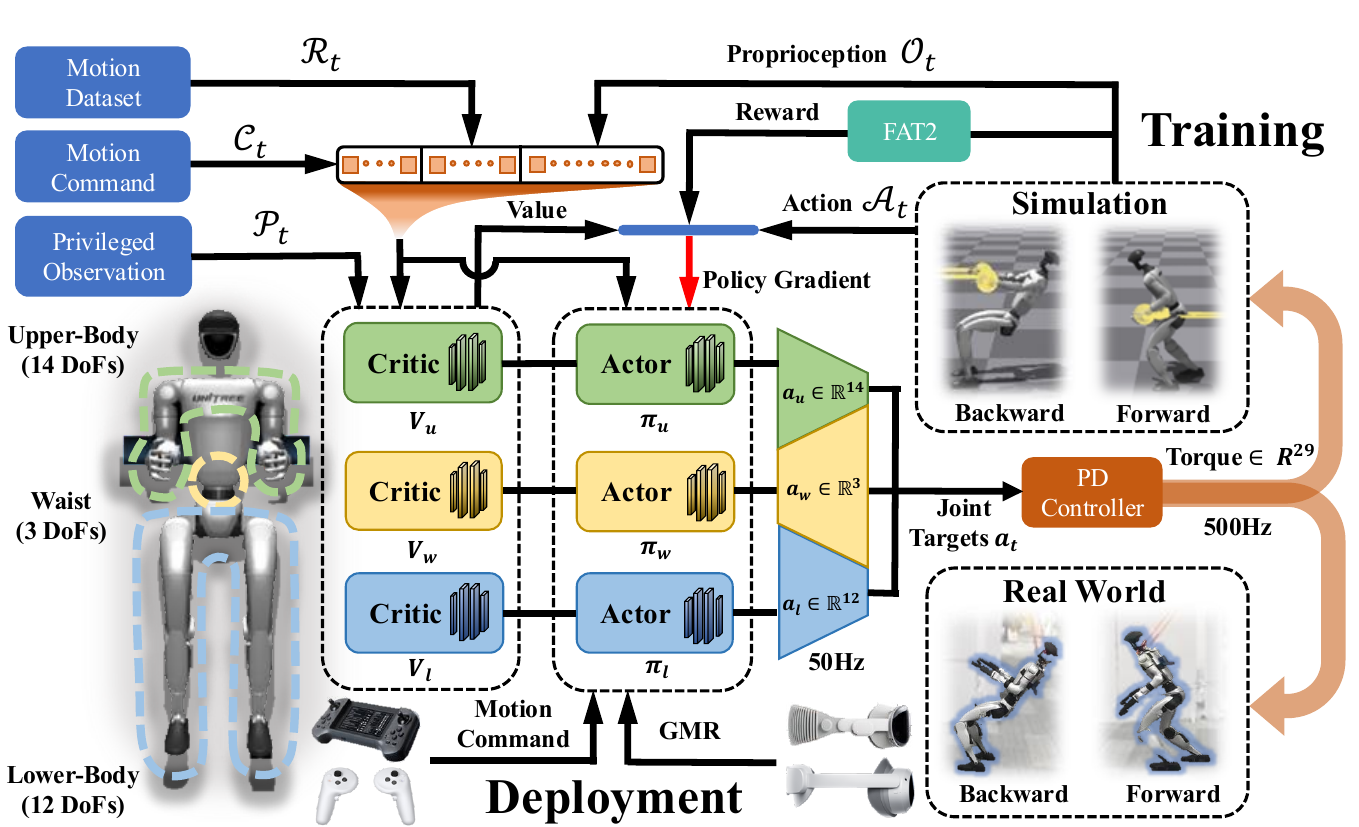}
\caption{
Pipeline of Thor. Three separate actor–critic networks control the upper body, waist, and lower body, with critics receiving privileged EE force information. FAT2 encourages force-adaptive postural adjustment. Upper-body references come from a human motion dataset during training and from VR-captured human motions retargeted to the robot using GMR during deployment. The deployed actors receive these references and remote-controller commands to output joint position targets, which PD controllers convert into joint torques.
}
\label{fig2}
\end{figure*}

Under large hand-contact forces, humanoids must adapt their
whole-body posture to counteract external moments while
tracking commanded waist and arm motions. This requires
coordinating load-dependent postural adaptation with
reference tracking across mechanically coupled body regions.

Building on FALCON~\cite{Falcon}, Thor explicitly separates
waist posture regulation from lower-body locomotion and
force adaptation, and from upper-body motion tracking.
The three policies share whole-body observations but optimize
body-specific objectives. To guide load-dependent postural
adaptation, we derive FAT2 from quasi-static moment balance,
which maps hand-contact forces and their moment arms to a
reference horizontal center-of-mass (CoM) offset. This
objective guides lower-body adaptation while the waist
and upper-body policies track their respective references.
Training uses a two-stage curriculum with increasing force disturbances and randomizes their magnitudes and directions at the end-effectors (EEs).
We evaluate Thor against baselines in simulation and on hardware, and demonstrate tasks including towing a 1.7-ton passenger car, opening a fire door, pushing a wheelchair, and wiping a whiteboard, as illustrated in Fig.~1.

The main contributions of this work are as follows:

    

\begin{itemize}
    \item We propose a body-part-factorized control
    architecture that separately optimizes lower-body
    locomotion and force adaptation, waist posture
    regulation, and upper-body motion tracking.
    Capacity-matched ablations demonstrate improved
    whole-body tracking under large external force
    disturbances.

    \item We formulate FAT2, a force-adaptive CoM offset
    objective derived from quasi-static moment balance.
    The reference accounts for hand-contact forces and
    their moment arms, providing a load-dependent
    postural objective that improves real-world peak
    pulling capability.

    \item We validate Thor on the Unitree G1 through
    quantitative pulling-force comparisons, component
    ablations, and real-world force-interaction tasks,
    demonstrating improved pulling capability and
    task performance under substantial external loads.
\end{itemize}

\section{Related Work}

\subsection{Forceful Interaction in Legged Robots}

Forceful legged manipulation requires maintaining whole-body stability while opening doors, carrying payloads, or pushing and pulling loads \cite{2016Humanoids_Motion_generation,2018TRO-Quadratic_Programming,2019RAL-Torque-Based_Balancing,2005ICRA}.
Model-based methods coordinate balance and task execution through explicit dynamics modeling, whole-body optimization, or force control \cite{2024TRO_Adaptive_Force_Based,2016RAL-Interaction_Force_Reconstruction,2023ICRA-Hierarchical_Adaptiv,2024ICRA-Hierarchical_Optimization-based,2025ICRA-Full-Order,HECTOR,Kinodynamics}.
However, their dependence on accurate models and contact or force information can limit adaptability to uncertain, dynamic interactions.

Learning-based methods offer alternative strategies for forceful interaction.
Portela et al. \cite{2024ICRA-Learning_Force_Contro} learn task-space force control for legged manipulation, while FACET \cite{FACET} achieves controllable compliance by tracking virtual mass-spring-damper dynamics.
FALCON \cite{Falcon} progressively increases end-effector forces during training to improve loco-manipulation robustness.
For humanoids under substantial loads, HAFO \cite{HAFO} combines virtual spring-damper disturbance modeling with curriculum learning, while FAME \cite{FAME} captures the coupling between upper-body configurations and bimanual forces to improve balance under uncertain hand loads.
HEFT \cite{HEFT} uses a motion-dependent payload curriculum to enable robust whole-body teleoperation.

These approaches mainly emphasize stability, interaction regulation, or load adaptation, with less attention to exploiting whole-body posture to increase sustainable interaction forces.
Thor addresses this through a force-adaptive torso-tilt (FAT2) reward derived from force analysis, encouraging load-dependent postural adjustment for intense force-interaction tasks.

\subsection{Policy Architectures for Humanoid Robots}

With the rapid progress of humanoid whole-body control, teleoperation, and loco-manipulation, how to structure high-dimensional whole-body policies has become an important design consideration.
One line of work adopts a unified whole-body policy to coordinate different body regions through shared representations.
OmniH2O \cite{2024CoRL-OmniH2O} learns a general-purpose humanoid controller through whole-body motion tracking, while HOVER \cite{2025ICRA-HOVER} distills multiple control modes into a unified policy.
CLONE \cite{Clone} further employs a Mixture-of-Experts architecture to improve coordination for complex whole-body motions.

Another line of work introduces functional decomposition to specialize different body regions or control objectives.
HOMIE \cite{2025RSS-HOMIE} separates robust lower-body locomotion from upper-body manipulation, while FALCON \cite{Falcon} and HAFO \cite{HAFO} adopt dual-policy architectures to coordinate lower-body stability and upper-body task execution.
More recent works introduce finer-grained structures at the objective and action levels.
CWI \cite{CWI} employs multiple critics to separately optimize locomotion, manipulation, and motion-style objectives, while HAF \cite{HAF} hierarchically organizes whole-body action generation over locomotion, waist posture, and bimanual manipulation.
Building on upper--lower body policy decomposition, Thor
separately optimizes waist posture regulation and lower-body
force adaptation, and introduces a moment-balance-derived
CoM objective for load-dependent postural adjustment.
We evaluate these design choices through capacity-matched
tracking comparisons and real-world pulling-force ablations.

\section{Methodology}

In force-interaction tasks, a key challenge for humanoids is coordinating locomotion, postural adaptation, and upper-body task execution while maintaining whole-body stability and sufficient force output at the end-effectors (EEs).
Thor addresses this challenge through an RL-based whole-body controller that combines body-part policy factorization with a force-adaptive torso-tilt objective to support coordinated responses to external loads, as shown in Fig.~\ref{fig2}. 

The WBC based on RL for humanoids can be modeled as a Markov Decision Process (MDP). The state space is defined as $\mathcal{S}_{t}=(\mathcal{O}_{t}, \mathcal{P}_{t}, \mathcal{A}_{t}, \mathcal{C}_{t}, \mathcal{R}_{t})$, where $\mathcal{O}_{t}=\left(q_{t}, \dot{q}_{t}, w_{t}, g_{t}\right)$ represents the robot’s own observations, including joint positions $q_{t} \in \mathbb{R}^{29}$, joint velocities $\dot{q}_{t} \in \mathbb{R}^{29}$, angular velocities $w_{t} \in \mathbb{R}^{3}$, and the projection of gravity onto the local coordinate frame $g_{t} \in \mathbb{R}^{3}$. The robot’s previous action output is defined as $\mathcal{A}_{t}=a_{t-1} \in \mathbb{R}^{29}$. The privileged information is $\mathcal{P}_{t}=\left(v_{t}, o_{t}, F_{t}\right)$, which includes the robot’s linear velocity $v_{t} \in \mathbb{R}^{3}$, orientation represented by a quaternion $o_{t} \in \mathbb{R}^{4}$, and external force $F_{t} \in \mathbb{R}^{6}$. 
The control commands are $\mathcal{C}_{t}=\left(v_{t}^{lin}, w_{t}^{ang}, \psi_{t}^{mode}, h_{t}^{root}, q_{t}^{w,ref}\right)$,
consisting of planar linear velocity
$v_{t}^{lin}\in\mathbb{R}^{2}$,
angular velocity about the z-axis
$w_{t}^{ang}\in\mathbb{R}^{1}$,
locomotion mode $\psi_{t}^{mode}\in\mathbb{R}^{1}$,
hip height $h_{t}^{root}\in\mathbb{R}^{1}$,
and target waist joint angles
$q_{t}^{w,ref}\in\mathbb{R}^{3}$ for roll, yaw, and pitch.
$\mathcal{R}_{t}=q_{t}^{ref}\in\mathbb{R}^{14}$
represents the target joint angles of the robot's upper body.

\subsection{Body-Part-Factorized Policy Architecture}

Load-dependent whole-body adaptation and waist reference
tracking impose distinct control objectives. The lower body
must adjust locomotion and posture in response to external
forces, while the waist regulates the relative orientation
between the trunk and pelvis. Motivated by these distinct
roles, Thor assigns waist regulation to a dedicated policy
with its own reward and value function. FAT2 guides the
lower-body policy, while the waist and upper-body policies
optimize their respective tracking objectives. All policies
share whole-body observations and are jointly trained
through the coupled robot dynamics. This design allows
separate optimization of these objectives while retaining
the information needed for coordinated control.

Specifically, we factorize the whole-body policy as $\pi=\left[\pi_l,\pi_w,\pi_u\right]$. The lower-body policy $\pi_l$ controls 12 joints across the hips, knees, and ankles of both legs for locomotion, balance, and force-adaptive postural adjustment. The waist policy $\pi_w$ controls the three waist joints (pitch, roll, and yaw) and tracks the reference waist posture to regulate the relative orientation between the trunk and pelvis. Within the jointly trained architecture, this posture-regulation objective is intended to support upper-body motion tracking while the lower body adapts to external forces. The upper-body policy $\pi_u$ controls 14 joints across both arms and tracks upper-body reference motions sampled from the AMASS dataset \cite{AMASS} during training.

\begin{figure}[!t]
\centering
\includegraphics[width=0.3\textwidth,]{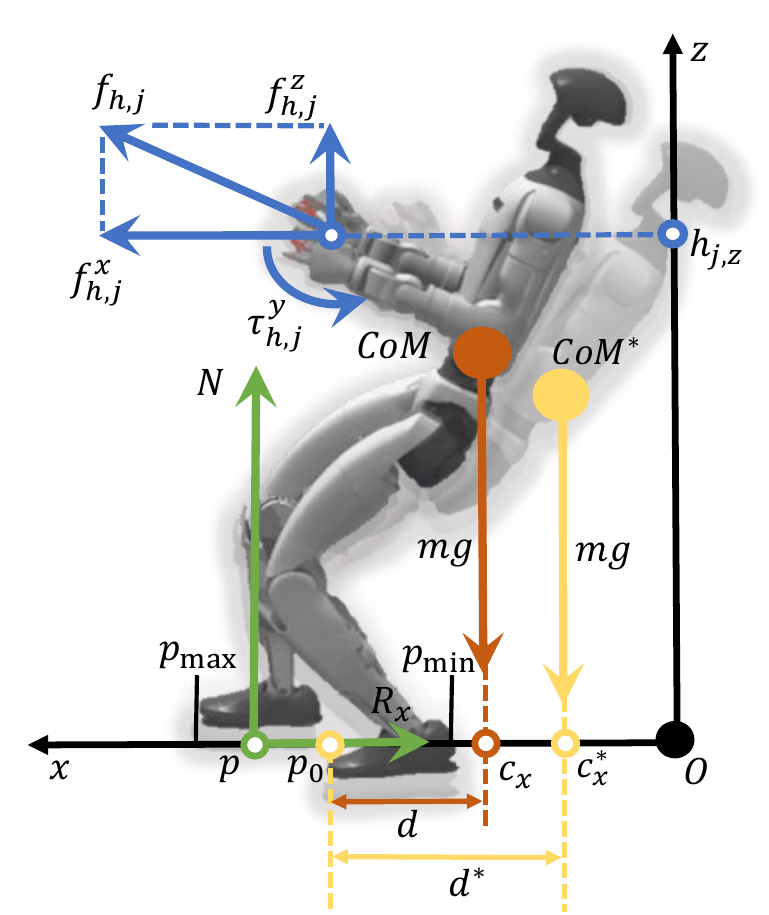}
\caption{
Humanoid force interaction analysis with ZMP constraint.
}
\label{fig3}
\end{figure}

The three agents share the same observation vector
$x_t=(\mathcal{O}_t,\mathcal{A}_t,\mathcal{C}_t,\mathcal{R}_t)$,
while maintaining separate actor and critic parameters.
For each body part $i\in\mathcal{I}=\{l,w,u\}$, the actor
$\pi_{\theta^i}(a_t^i\mid x_t)$ outputs actions for its assigned joints.
The corresponding critic $V_{\phi^i}(s_t)$ additionally receives
privileged information $\mathcal{P}_t$, where $s_t=(x_t,\mathcal{P}_t)$,
for value estimation during training.
The joint policy is factorized over the three action groups:
\begin{equation}
\pi_{\Theta}(a_t\mid x_t)
=
\prod_{i\in\mathcal{I}}
\pi_{\theta^i}(a_t^i\mid x_t),
\end{equation}
where $\Theta=\{\theta^i\}_{i\in\mathcal{I}}$, and the component
actions are assembled into the whole-body action $a_t$ according
to their joint assignments.

The policies collect trajectories jointly in the same environment,
but each uses its own reward $r_t^i$ and critic to compute
generalized advantage estimates.
For a trajectory segment ending at $T$,
\begin{equation}
\begin{gathered}
\delta_t^i
=r_t^i+\gamma V_{\phi^i}(s_{t+1})
-V_{\phi^i}(s_t),\\
\hat{A}_t^i
=\sum_{k=0}^{T-t-1}(\gamma\lambda)^k\delta_{t+k}^i.
\end{gathered}
\end{equation}
where $\gamma$ is the discount factor and $\lambda$ is the GAE
parameter.
Bootstrap values are set to zero at terminal states and estimated
by the corresponding critic at nonterminal segment boundaries.

Each actor--critic pair is updated using PPO \cite{PPO}, with its
own action probability ratios and advantage estimates.
The combined PPO objective is
\begin{equation}
\mathcal{L}_{\mathrm{PPO}}(\Theta,\Phi)
=
\sum_{i\in\mathcal{I}}
\left[
L_i^{\mathrm{CLIP}}
-c_v L_i^{\mathrm{VF}}
+c_e L_i^{\mathrm{S}}
\right],
\end{equation}
where $\Phi=\{\phi^i\}_{i\in\mathcal{I}}$, and
$L_i^{\mathrm{CLIP}}$, $L_i^{\mathrm{VF}}$, and $L_i^{\mathrm{S}}$
denote the standard PPO clipped surrogate objective, value
regression loss, and policy entropy, respectively.
The coefficients $c_v$ and $c_e$ weight the value loss and
entropy term.

To accelerate policy convergence, we used a two-stage curriculum learning approach based on task difficulty. Initially, the robot was trained in a low-force disturbance environment to build stable motion capabilities. Subsequently, environmental difficulty was increased with extreme force disturbances to improve the humanoid robot’s force-interaction task performance. To bridge the sim-to-real gap, domain randomization following a Gaussian distribution was applied to the direction and magnitude of forces on the robot’s EEs, better simulating real-world contact-rich force-interaction environments.

\subsection{ZMP-Inspired Force-Adaptive Torso-Tilt Reward}

FAT2 promotes load-dependent postural adaptation through a horizontal
center-of-mass (CoM) offset objective. We derive the
reference from sagittal-plane quasi-static equilibrium, neglecting CoM
acceleration and the rate of change of centroidal angular momentum.

Consider coplanar foot contacts on the horizontal plane $z=0$, with $x$
horizontal and $z$ upward. Let $m$ denote the robot mass, $g>0$ the
gravitational acceleration, and $c_x$ the whole-body CoM coordinate.
At hand position $\mathbf h_j$, the environment exerts a force
$\mathbf f_{h,j}$ and a contact moment $\boldsymbol\tau_{h,j}$ on the
robot, all expressed in the world frame; $\boldsymbol\tau_{h,j}=\mathbf 0$
for point-force disturbances.
The resultant normal and horizontal ground reaction forces, $N$ and
$R_x$, satisfy
\begin{equation}
N=mg-\sum_j f_{h,j}^{z},
\qquad
R_x=-\sum_j f_{h,j}^{x}.
\label{eq:fat2_force_balance}
\end{equation}

Let $p$ denote the sagittal coordinate of the zero-moment point
(ZMP) associated with the resultant foot--ground contact wrench.
The interval $[p_{\min},p_{\max}]$ is the sagittal projection
of the active foot-support polygon.
We update $p_0=(p_{L,x}+p_{R,x})/2$ during double support,
where $p_{L,x}$ and $p_{R,x}$ are the sagittal foot-center
coordinates, and hold it fixed during single support
to avoid changes caused by swing-foot motion.
During single support, \(p_0\) may lie outside the support interval; FAT2 then serves as a posture prior without guaranteeing quasi-static feasibility.
Moment equilibrium about the ground point $(p_0,0)$ gives
\begin{equation}
\begin{gathered}
M_h(p_0)=\sum_j\Bigl[
h_{j,z}f_{h,j}^{x}
-(h_{j,x}-p_0)f_{h,j}^{z}
+\tau_{h,j}^{y}\Bigr],\\
N(p-p_0)=mg(c_x-p_0)+M_h(p_0),
\end{gathered}
\label{eq:fat2_moment_balance}
\end{equation}
where $M_h(p_0)$ is the resultant hand-contact pitch moment about this
point.

The desired condition $p=p_0$ defines the reference CoM offset $d^*$.
Its tracking error $e_d$ satisfies
\begin{equation}
\begin{gathered}
d^*=-\frac{M_h(p_0)}{mg},\\
e_d=(c_x-p_0)-d^*
=\frac{N}{mg}(p-p_0).
\end{gathered}
\label{eq:fat2_com_reference}
\end{equation}
Thus, for $N>0$, zero offset error corresponds to $p=p_0$ under the
quasi-static assumptions.
We compute $c_x$ from the mass-weighted link CoM positions and update
$d^*$ using the current hand positions and interaction wrenches.
This objective accommodates different whole-body configurations without
prescribing a unique torso pitch.

\begin{figure*}[!t]
\centering
\includegraphics[width=0.9\textwidth,]{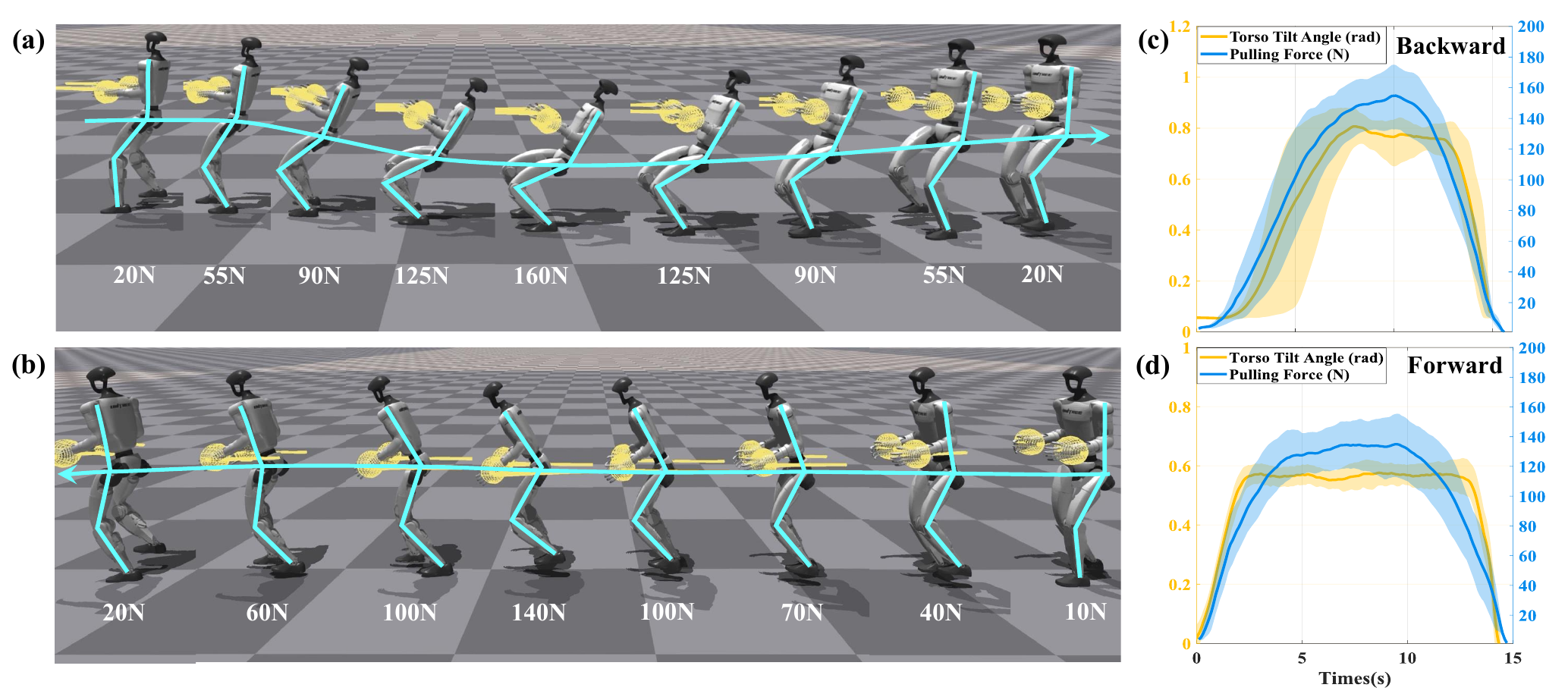}
\caption{
Force-dependent postural adaptation in simulation and the real world.
(a,b) Simulated posture sequences and corresponding interaction forces during backward and forward locomotion, respectively.
(c,d) Real-world time histories of pulling force and torso tilt angle during backward and forward locomotion, respectively.
}
\label{fig:simulated posture sequences}
\end{figure*}

The FAT2 reward is defined as
\begin{equation}
r_{\mathrm{FAT2}}
=\exp\!\left[-\frac{(d-d^*)^2}{s_d^2}\right],
\qquad d=c_x-p_0,
\label{eq:fat2_reward}
\end{equation}
where $s_d>0$ is the distance-error scale in metres.
The reward is assigned to the lower-body policy with weight $2.0$.
Feasible foot contacts require $N>0$, $p\in[p_{\min},p_{\max}]$, and
admissible friction forces~\cite{Caron2016ZMPSupportAreas}.
The quasi-static solution serves as a physically grounded nominal posture reference rather than a hard constraint on robot motion. During full-dynamics training, FAT2 guides policy exploration toward force-adaptive postures as a soft reward, facilitating policy learning while allowing proprioceptive feedback to accommodate transient dynamics.

The main reward terms and their weights are provided
in Table~\ref{tab:reward_terms_by_policy} in the Appendix.

\section{Experiments and Results}
In this section, we conduct both simulation and real-world experiments on Unitree’s robot G1 to quantitatively evaluate the performance of Thor against baseline algorithms in force-interaction tasks, thereby validating the effectiveness of our approach.
G1 has 29 DoFs (12 in the lower body, 3 in the waist, and 14 in the upper body), with a total height of 1.32 m and a weight of 35 kg. This section primarily addresses the following three questions:

\begin{itemize}
    \item \textbf{Q1:}
    What are the effects of FAT2 and body-part policy factorization on force capability and whole-body coordination?
    \item \textbf{Q2:}
    How much does Thor improve force-interaction performance over baseline methods?
    \item \textbf{Q3:}
    How does Thor compare with baseline methods on force-interaction tasks in complex real-world environments?
\end{itemize}

\subsection{Simulation Experiments}

We conducted the RL policy training of Thor in the simulator Isaac Gym.
All training was performed on an NVIDIA RTX 4090 GPU, with each curriculum learning stage lasting approximately 3.4 hours. The main hyperparameters used in training are summarized in \textbf{Table~\ref{tab:Hyperparameters}}.

\begin{table}[t]
    \centering
    \caption{Main Hyperparameters}
    \label{tab:Hyperparameters}

    \fontsize{7.5pt}{9pt}\selectfont
    \setlength{\tabcolsep}{2pt}
    \renewcommand{\arraystretch}{1.2}

    \begin{tabularx}{\columnwidth}{
    @{}
    >{\hsize=0.9\hsize
      \linewidth=\hsize
      \raggedright\arraybackslash}X c
    @{\hspace{8pt}}
    >{\hsize=1.1\hsize
      \linewidth=\hsize
      \raggedright\arraybackslash}X c
    @{}
}
        \toprule
        \textbf{Hyperparameter} & \textbf{Value}
        & \textbf{Hyperparameter} & \textbf{Value} \\
        \midrule

        Parallel Environments & 4096
        & Value Loss Coefficient $c_v$ & 0.9 \\

        Training Iterations & $1 \times 10^{4}$
        & GAE Lambda $\lambda$ & 0.95 \\

        Hidden Layers & [512, 256, 128]
        & Target KL Divergence & 0.01 \\

        Learning Rate & $5 \times 10^{-4}$
        & Rollout Horizon $T$ & 24 \\

        Discount Factor $\gamma$ & 0.98
        & Swing-Foot Height Target $z^\star$
        & $0.12\,\mathrm{m}$ \\

        Epsilon Clip $\varepsilon_{\mathrm{clip}}$ & 0.15
        & Entropy Coefficient $c_e$ & 0.02 \\

        \bottomrule
    \end{tabularx}
\end{table}

To address \textbf{Q1}, we first evaluated in the simulation environment how the robot’s posture changes with varying pulling forces.
As shown in Fig.~\ref{fig:simulated posture sequences}(a,b), the robot remains nearly upright under small pulling forces. As the force increases during backward or forward locomotion, it progressively leans to counteract the external load. At high loads, the ground projection of its CoM lies outside the foot-support polygon. As the force decreases, the robot gradually returns to an upright posture.
These observations illustrate the load-dependent postural adaptation of Thor, consistent with the intended role of FAT2 in encouraging the robot to adjust its posture in response to external forces.

To further evaluate body-part policy factorization, we conduct simulation ablations comparing Thor with two-policy and single-policy variants. In the two-policy variant, one policy jointly controls the lower body and waist, while the other controls the upper body. The single-policy variant controls all joints using one policy. All variants retain the same 29 controllable DoFs and FAT2, and are evaluated using the same external-force profiles and command sequences.
To control for network capacity, we adjusted the hidden-layer widths such that the aggregate actor and critic parameter counts of the single-, two-, and three-policy variants differ by less than 1\%. All variants were trained from scratch using the same observation and action spaces, PPO configuration, environment-interaction budget, and number of training iterations. This capacity-matched design prevents additional model capacity from confounding the comparison of policy factorization.
The single-policy variant uses $r_t^l+r_t^w+r_t^u$;
the two-policy variant uses $r_t^l+r_t^w$ and $r_t^u$
for its lower-body--waist and upper-body policies, respectively.
Each policy has its own critic, with identical GAE
and advantage-normalization settings across variants.


\begin{figure*}[!t]
\centering
\includegraphics[width=0.8\textwidth,]{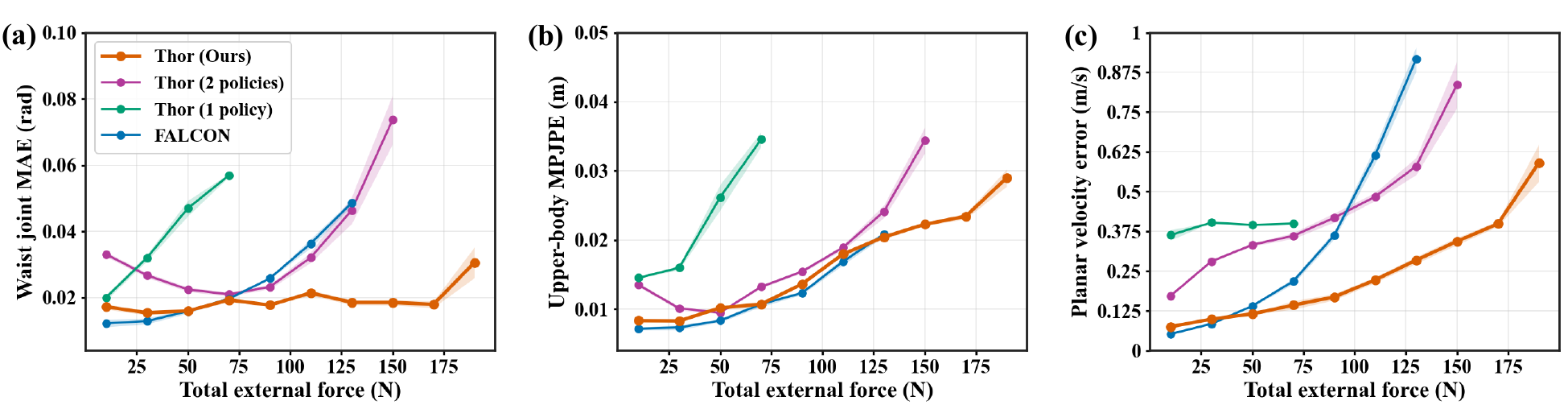}
\caption{Whole-body tracking performance under external
force disturbances, comparing Thor, its single- and
two-policy variants, and FALCON:
(a) waist joint mean absolute error;
(b) upper-body mean per-joint position error;
(c) planar velocity tracking error.
Lower values indicate better performance.
Curves show the mean across independent evaluation trials
satisfying the validity criteria within each force interval,
and shaded bands indicate $\pm$SEM.
Each trial uses a different evaluation random seed.}
\label{fig:policy_factorization_ablation}
\end{figure*}

We evaluate whole-body tracking using waist joint mean
absolute error (MAE), upper-body mean per-joint position
error (MPJPE), and planar velocity tracking error, with
FALCON included as an additional baseline.
Tracking errors are evaluated only under non-falling conditions.
Curves terminate at lower force levels when the corresponding
method experiences a fall or tracking failure.
As shown in Fig.~\ref{fig:policy_factorization_ablation}(a),
Thor maintains low waist tracking errors as the external
force increases, whereas the errors of the single- and
two-policy variants rise substantially under larger loads.
Fig.~\ref{fig:policy_factorization_ablation}(b,c) further
shows that Thor exhibits smaller increases in upper-body
position and planar velocity errors, outperforming both
ablation variants and FALCON under larger external forces.
Together, these results indicate that Thor better preserves
posture regulation, upper-body motion tracking, and
locomotion tracking during forceful interactions.

The ablations in Fig.~\ref{fig:policy_factorization_ablation}
specifically evaluate the contribution of body-part policy
factorization to tracking accuracy under large external
force disturbances. All three Thor variants retain FAT2
and the same controllable DoFs, with matched model capacity
and training budgets. The lower tracking errors achieved
by the three-policy architecture therefore support the
benefit of body-part policy factorization with body-specific
objectives. In particular, the comparison with the two-policy
variant shows that assigning waist regulation to a dedicated
policy improves waist and upper-body tracking while also
reducing locomotion tracking errors under large external
loads.

\subsection{Real-World Evaluation}

We deployed Thor on the G1 robot with the three actor networks evaluated at 50 $\mathrm{Hz}$. 
Their outputs were assembled according to the joint assignments to form
whole-body joint-position targets. 
A PD controller computed the joint torques, which were transmitted to the motors at 500 $\mathrm{Hz}$.
Fig.~\ref{fig:simulated posture sequences}(c,d) shows pulling
force and torso tilt during backward and forward locomotion,
respectively. Curves and shaded bands represent the pointwise
mean and $\pm$SEM across repeated real-world trials.
The robot exhibits load-dependent torso tilting, consistent with the intended role of FAT2 in encouraging postural adaptation under external forces.
To further address \textbf{Q1}, we compare Thor with
$\mathrm{Thor}^{1}$, which retains FAT2 with two policies,
and $\mathrm{Thor}^{2}$, which uses three policies without
FAT2. Table~\ref{tab:Experimental data} shows that adding
FAT2 to $\mathrm{Thor}^{2}$ increases mean peak dual-hand
pulling force by $60.3\%$ and $40.4\%$ during backward
and forward locomotion, respectively. With FAT2 retained,
separating waist control into a dedicated policy
(Thor vs.\ $\mathrm{Thor}^{1}$) yields corresponding gains
of $21.2\%$ and $13.7\%$. These results demonstrate the
contributions of both components to peak pulling capability.

To answer \textbf{Q2}, we evaluated Thor and the baseline methods under multiple conditions by measuring the mean peak pulling forces. 
During the pulling-force tests, the rope was kept horizontal and at the same height as the robot's hand end effectors.
Under each test condition, all methods used identical upper-body reference postures and lower-body motion commands.
For each method and condition, we conducted 20 trials and extracted the peak pulling force from a continuous 10 $\mathrm{s}$ interval of stable pulling, excluding
transient peaks caused by rope take-up, abrupt acceleration, or instability.
\textbf{Table~\ref{tab:Experimental data}} reports the mean and standard error of these trial-level peaks.
The baseline methods include Falcon \cite{Falcon}, HEFT \cite{HEFT}, SONIC \cite{SONIC}, and Homie \cite{2025RSS-HOMIE}.
All baselines use official checkpoints and deployment settings,
with load-trained variants when available.
Under identical test conditions, functionally equivalent commands
are supplied through each method's native interface,
using the formats and ranges employed during training.
These conditions include the robot operating with a single-hand $(s)$, dual-hand $(d)$, forward $(f)$, backward $(b)$, and standing in place $(p)$, as well as generating pulling forces in different directions. 
For example, $F_{db}^{180^{\circ}}$ denotes the pulling force produced by the robot when using both hands during backward locomotion, directed at $180^{\circ}$ relative to its positive x-axis.

\begin{table*}
\vspace{0.2cm}
\centering
\caption{Experimental data of Thor and baselines (Mean $\pm$ SE; larger is better)}
\label{tab:Experimental data}
\begin{tblr}{
  rowsep = 2pt, 
  colspec = {p{0.8cm}p{0.7cm}p{1.0cm}p{0.9cm}p{0.9cm}p{1.0cm}p{1.0cm}p{1.0cm}p{1.0cm}p{1.0cm}p{0.9cm}p{1.0cm}p{0.9cm}},
  cells = {mode=math, c}, 
  cell{2}{1} = {r=5}{},
  cell{7}{1} = {r=2}{},
  hline{1,9} = {-}{0.12em},
  hline{2,7} = {-}{0.08em},
}

\text{Category} & \text{Method} &
F_{db}^{180^{\circ}}(\text{N})\!\uparrow &
F_{df}^{0^{\circ}}(\text{N})\!\uparrow &
F_{sf}^{0^{\circ}}(\text{N})\!\uparrow &
F_{sb}^{180^{\circ}}(\text{N})\!\uparrow &
F_{dp}^{180^{\circ}}(\text{N})\!\uparrow &
F_{dp}^{135^{\circ}}(\text{N})\!\uparrow &
F_{dp}^{90^{\circ}}(\text{N})\!\uparrow &
F_{dp}^{45^{\circ}}(\text{N})\!\uparrow &
F_{dp}^{0^{\circ}}(\text{N})\!\uparrow &
F_{sp}^{180^{\circ}}(\text{N})\!\uparrow &
F_{sp}^{0^{\circ}}(\text{N})\!\uparrow \\

\text{Data} & \textbf{Thor} & \mathbf{167.7_{\pm2.4}} & \mathbf{145.5_{\pm2.0}} & \mathbf{78.2_{\pm5.0}} &
\mathbf{147.5_{\pm4.9}} & \mathbf{59.9_{\pm1.1}} & \mathbf{127.4_{\pm3.3}} &
\mathbf{73.6_{\pm2.0}} & \mathbf{59.2_{\pm2.3}} & \mathbf{35.4_{\pm0.4}} &
\mathbf{64.9_{\pm2.0}} & \mathbf{58.5_{\pm3.6}} \\

& \text{Falcon} & 99.3_{\pm1.3} & 83.3_{\pm2.7} & \mathbf{78.1_{\pm4.5}} &
92.4_{\pm6.2} & 53.4_{\pm1.3} & 97.8_{\pm3.2} &
66.2_{\pm2.9} & 30.0_{\pm0.43} & 27.1_{\pm2.4} &
55.7_{\pm2.3} & 29.4_{\pm1.4} \\

& \text{HEFT} &
91.9_{\pm13.1} & 82.6_{\pm4.2} & 51.0_{\pm1.0} &
79.8_{\pm4.9} & \mathbf{56.8_{\pm4.8}} & 54.0_{\pm4.6} &
59.4_{\pm5.6} & 36.1_{\pm1.7} & \mathbf{34.8_{\pm2.9}} &
53.5_{\pm2.7} & 45.0_{\pm22.1} \\

& \text{SONIC} &
55.4_{\pm1.3} & 44.2_{\pm4.3} & 44.3_{\pm6.7} &
41.5_{\pm8.1} & 39.4_{\pm4.1} & 44.4_{\pm3.4} &
42.5_{\pm2.3} & 27.9_{\pm3.1} & 31.0_{\pm2.7} &
24.9_{\pm2.6} & 31.7_{\pm1.1} \\

& \text{Homie} & 62.3_{\pm3.7} & 48.1_{\pm2.9} & 59.6_{\pm1.5} &
51.8_{\pm5.4} & 46.0_{\pm1.2} & 40.7_{\pm2.3} &
33.9_{\pm0.8} & 38.5_{\pm1.9} & \mathbf{34.2_{\pm2.3}} &
44.3_{\pm1.1} & 35.0_{\pm2.0} \\

\text{Ablation} & \text{Thor}^{1} & 138.4_{\pm5.4} & 128.0_{\pm3.5} &
72.5_{\pm2.7} & 104.6_{\pm2.4} & 54.9_{\pm2.4} &
103.5_{\pm1.5} & 68.0_{\pm0.9} & 41.2_{\pm0.8} &
33.0_{\pm1.5} & \mathbf{61.5_{\pm3.4}} & 44.6_{\pm2.3} \\

& \text{Thor}^{2} & 104.6_{\pm4.6} & 103.6_{\pm3.9} &
70.3_{\pm3.0} & 98.5_{\pm2.1} & 54.1_{\pm2.4} &
98.4_{\pm1.6} & 67.1_{\pm0.7} & 43.9_{\pm2.0} &
30.9_{\pm1.5} & 57.9_{\pm1.3} & 43.7_{\pm2.3} \\

\end{tblr}
\end{table*}

The experimental results demonstrate that Thor outperforms the baseline methods in most force-interaction tasks. 
Specifically, G1 achieved mean peak pulling force of 167.7 $\pm$ 2.4 $N$ (approximately 48.8\% of its body weight) during backward locomotion with dual-hand, and 145.5 $\pm$ 2.0 $N$ during forward locomotion.
Compared with the best-performing baseline method, Falcon, these results represent improvements of  68.9\% and 74.7\%, respectively.
Thor also improves single-hand backward pulling by 59.6\% over Falcon, while single-hand forward performance remains comparable. While standing in place, Thor achieves higher mean peak forces than the best baseline in six of seven tested configurations, reaching 127.4 $\pm$ 3.3 $N$ in dual-hand pulling at $135^\circ$. 
Together, Table~\ref{tab:Experimental data} and
Fig.~\ref{fig:policy_factorization_ablation} demonstrate
two complementary benefits of Thor: FAT2 substantially
enhances peak pulling capability, while the three-policy
architecture improves whole-body tracking accuracy under
large external force disturbances. The real-world ablations
additionally show that the dedicated waist policy further
improves pulling capability when FAT2 is retained.








To answer \textbf{Q3}, we evaluated Thor and the baselines on four real-world force-interaction tasks, conducting 20 trials per method per task to measure success rates. As shown in \textbf{Table~\ref{tab:real_world_tasks}}, Thor achieved higher success rates than all baselines across all four tasks.
In all manipulation tasks, the operator's upper-body motions were captured using VR and retargeted to the robot's upper body using General Motion Retargeting (GMR)~\cite{GMR} for teleoperation.
For fire-door opening, the robot hooked its end effector onto the handle and maintained balance while applying over 60 $\mathrm{N}$ with one hand and stepping backward to open the door before passing through it.
Failure cases among the baseline methods were primarily associated with instability under one-handed loading.
For wheelchair pushing and water lifting, teleoperation guided the Dex3-1 hands to grasp the wheelchair handles or pass through the water bag’s handle openings. 
Thor then maneuvered the wheelchair carrying a 60 $\mathrm{kg}$ payload and lifted the 10 $\mathrm{kg}$ water bag, demonstrating its ability to combine task-relevant motion tracking with substantial force interaction.
In the car-pulling task, it towed a $1.7$-ton car, placed in neutral with two adults inside, over 5 $\mathrm{m}$, generating approximately 160 $\mathrm{N}$ of pulling force during backward locomotion. The baselines failed because they could not generate sufficient backward pulling force to move the car. By overcoming the vehicle’s initial resistance to motion and sustaining traction throughout the pull, Thor demonstrated its capacity for sustained whole-body pulling in demanding real-world conditions.

\begin{table}[!t]
    \centering
    \caption{Success rates (\%) in real-world force-interaction scenarios.}
    \label{tab:real_world_tasks}

    \footnotesize
    \setlength{\tabcolsep}{3pt}
    \renewcommand{\arraystretch}{1.2}

    \begin{tabular*}{\columnwidth}{
        @{\extracolsep{\fill}}lcccc@{}
    }
        \toprule
        Method
        & \makecell{Water lifting\\($10\,\mathrm{kg}$)}
        & \makecell{Wheelchair\\pushing}
        & \makecell{Fire-door\\opening}
        & \makecell{Car pulling\\($1.7\,\mathrm{t}$)} \\
        \midrule

        \textbf{Thor}
        & \textbf{100} & \textbf{95} & \textbf{85} & \textbf{80} \\

        FALCON
        & 90 & 85 & 60 & 0 \\

        HEFT
        & 95 & 85 & 15 & 0 \\

        SONIC
        & 75 & 20 & 0 & 0 \\

        \bottomrule
    \end{tabular*}

    \vspace{2pt}
\end{table}


\section{Conclusions and Future Work}
In this paper, we presented Thor, a humanoid whole-body control framework for intense, contact-rich environments. FAT2 encourages force-dependent postural adaptation, while the body-part-factorized architecture promotes functional specialization and helps reduce interference between lower-body adaptation and upper-body task execution through waist coordination. Simulation and real-world experiments demonstrate improved pulling capability and coordinated task execution. By enabling physically demanding load-handling and towing tasks, Thor expands the practical capabilities of humanoids and brings them closer to productive work in factories and other industrial environments.

Our evaluation primarily covers short-duration force-interaction
tasks. Sustained material handling under varying ground friction
and interaction loads remains to be systematically evaluated.
Future work will investigate long-duration operation in industrial
environments, including actuator thermal limits and the consistency
of locomotion and manipulation performance over repeated task
execution.




\section*{APPENDIX}

\begin{table}[H]
\centering
\begingroup

\caption{Main reward terms and weights by policy.}
\label{tab:reward_terms_by_policy}

\fontsize{6.3pt}{7.1pt}\selectfont
\setlength{\tabcolsep}{2.0pt}
\renewcommand{\arraystretch}{1.0}

\begin{tabularx}{\columnwidth}{
  @{}X@{\hspace{15pt}}X@{}
}
\toprule
\textbf{Term}\hfill\textbf{Weight}
&
\textbf{Term}\hfill\textbf{Weight}
\\

\midrule
\multicolumn{2}{c}{%
  \hspace*{-0.4cm}%
  \textit{Common regularizers (per policy) \hspace{1.8cm} Lower-body policy}%
} \\
\midrule

\rewardcell
  {Action rate}
  {$-0.5$}
  {r_{\Delta a}^{k}
     =\lVert\mathbf a_t^k-\mathbf a_{t-1}^k\rVert_2^2}
&
\rewardcell
  {Command tracking}
  {$2.0,\;1.5,\;3.0$}
  {r_{v_i}=\exp\!\left(-\frac{e_{v_i}^{2}}{0.25}\right),
   \quad i=x,y,\omega}
\\
\addlinespace[0.65ex]

\rewardcell
  {Joint torque}
  {$-10^{-5}$}
  {r_{\tau}^{k}
     =\eta_k\lVert\boldsymbol{\tau}_k\rVert_2^2}
&
\rewardcell
  {Swing-foot height}
  {$-15.0$}
  {r_z=\sum_f(1-c_f)(z_f-z^\star)^2}
\\
\addlinespace[0.65ex]

\rewardcell
  {Joint velocity}
  {$-10^{-3}$}
  {r_{\dot q}^{k}
     =\lVert\dot{\mathbf q}_k\rVert_2^2}
&
\rewardcell
  {Feet air time}
  {$5.0$}
  {r_{\mathrm{air}}
     =I_v\sum_f(T_f-0.5)d_f}
\\
\addlinespace[0.65ex]

\rewardcell
  {Joint acceleration}
  {$-2.0{\times}10^{-7}$}
  {r_{\ddot q}^{k}
     =\left\lVert
       \frac{\Delta\dot{\mathbf q}_k}{\Delta t}
      \right\rVert_2^2}
&
\rewardcell
  {Force-adaptive torso-tilt}
  {$2.0$}
  {r_{\mathrm{FAT2}}=\exp\!\left[
     -\frac{(d-d^*)^2}{0.0225}
   \right]}
\\
\addlinespace[0.65ex]

\rewardcell
  {Joint-position limits}
  {$-5.0$}
  {r_{\mathrm{lim}}^{k}
     =\sum_{i\in k}
       \left(
         [q_i^- -q_i]_+
         +[q_i-q_i^+]_+
       \right)}
&
\rewardcell
  {Failure termination}
  {$-450.0$}
  {r_{\mathrm{term}}
     =\mathbb{I}_{\mathrm{reset}\,\wedge\,
       \neg\mathrm{timeout}}}

\\

\midrule
\multicolumn{2}{c}{\textit{Waist policy   \hspace{2.8cm}  Upper-body policy}} \\
\midrule

\rewardcell
  {Waist pose}
  {$3.0$}
  {r_q^{w}
     =\exp\!\left(
       -\frac{\lVert\mathbf e_w\rVert_2^2}{0.05}
     \right)}
&
\rewardcell
  {Upper-body pose tracking}
  {$4.0$}
  {r_q^u=\exp\!\left(
     -\frac{\lVert\mathbf e_u\rVert_2^2}{0.1}
   \right)}

\\
\addlinespace[0.65ex]

\rewardcell
  {Torso-pitch damping}
  {$-1.0$}
  {r_{\omega}^{w}
     =\,\omega_{y}^2}
&
\rewardcell
  {EE velocity smoothness}
  {$-0.1$}
  {r_{\Delta v}^{u}
     =\sum_h\lVert\Delta\mathbf v_h\rVert_2}
\\

\bottomrule
\end{tabularx}

\vspace{0.5ex}
\begin{minipage}{\columnwidth}
\fontsize{5.7pt}{6.5pt}\selectfont
\raggedright

$k\in\{l,w,u\}$ indexes the lower-body, waist, and
upper-body policies, respectively.
All reward weights are scaled by $\Delta t=0.02\,\mathrm{s}$.

\end{minipage}

\endgroup
\end{table}




\bibliographystyle{IEEEtran}
\bibliography{Mybib}

@misc{Hold_My_Beer,
      title={Hold My Beer: Learning Gentle Humanoid Locomotion and End-Effector Stabilization Control}, 
      author={Yitang Li and Yuanhang Zhang and Wenli Xiao and Chaoyi Pan and Haoyang Weng and Guanqi He and Tairan He and Guanya Shi},
      year={2025},
      eprint={2505.24198},
      archivePrefix={arXiv},
      primaryClass={cs.RO},
      url={https://arxiv.org/abs/2505.24198}, 
}

@misc{Opt2Skill_Imitating,
      title={Opt2Skill: Imitating Dynamically-feasible Whole-Body Trajectories for Versatile Humanoid Loco-Manipulation}, 
      author={Fukang Liu and Zhaoyuan Gu and Yilin Cai and Ziyi Zhou and Hyunyoung Jung and Jaehwi Jang and Shijie Zhao and Sehoon Ha and Yue Chen and Danfei Xu and Ye Zhao},
      year={2025},
      eprint={2409.20514},
      archivePrefix={arXiv},
      primaryClass={cs.RO},
      url={https://arxiv.org/abs/2409.20514}, 
}

@INPROCEEDINGS{2016Humanoids_Motion_generation,
  author={Ramirez-Alpizar, Ixchel G. and Naveau, Maximilien and Benazeth, Christophe and Stasse, Olivier and Laumond, Jean-Paul and Harada, Kensuke and Yoshida, Eiichi},
  booktitle={2016 IEEE-RAS 16th International Conference on Humanoid Robots (Humanoids)}, 
  title={Motion generation for pulling a fire hose by a humanoid robot}, 
  year={2016},
  volume={},
  number={},
  pages={1016-1021},
  doi={10.1109/HUMANOIDS.2016.7803396}}

@ARTICLE{2024TRO_Adaptive_Force_Based,
  author={Sombolestan, Mohsen and Nguyen, Quan},
  journal={IEEE Transactions on Robotics}, 
  title={Adaptive-Force-Based Control of Dynamic Legged Locomotion Over Uneven Terrain}, 
  year={2024},
  volume={40},
  number={},
  pages={2462-2477},
  doi={10.1109/TRO.2024.3381554}}

@ARTICLE{2021RAL_Humanoid_Loco-Manipulations,
  author={Murooka, Masaki and Chappellet, Kevin and Tanguy, Arnaud and Benallegue, Mehdi and Kumagai, Iori and Morisawa, Mitsuharu and Kanehiro, Fumio and Kheddar, Abderrahmane},
  journal={IEEE Robotics and Automation Letters}, 
  title={Humanoid Loco-Manipulations Pattern Generation and Stabilization Control}, 
  year={2021},
  volume={6},
  number={3},
  pages={5597-5604},
  doi={10.1109/LRA.2021.3077858}}

@ARTICLE{2016RAL-Interaction_Force_Reconstruction,
  author={Mattioli, Tommaso and Vendittelli, Marilena},
  journal={IEEE Robotics and Automation Letters}, 
  title={Interaction Force Reconstruction for Humanoid Robots}, 
  year={2017},
  volume={2},
  number={1},
  pages={282-289},
  doi={10.1109/LRA.2016.2601345}}

@ARTICLE{2018TRO-Quadratic_Programming,
  author={Bouyarmane, Karim and Chappellet, Kevin and Vaillant, Joris and Kheddar, Abderrahmane},
  journal={IEEE Transactions on Robotics}, 
  title={Quadratic Programming for Multirobot and Task-Space Force Control}, 
  year={2019},
  volume={35},
  number={1},
  pages={64-77},
  doi={10.1109/TRO.2018.2876782}}

@ARTICLE{2019RAL-Torque-Based_Balancing,
  author={Abi-Farraj, Firas and Henze, Bernd and Ott, Christian and Giordano, Paolo Robuffo and Roa, Máximo A.},
  journal={IEEE Robotics and Automation Letters}, 
  title={Torque-Based Balancing for a Humanoid Robot Performing High-Force Interaction Tasks}, 
  year={2019},
  volume={4},
  number={2},
  pages={2023-2030},
  doi={10.1109/LRA.2019.2898041}}

@misc{Falcon,
      title={FALCON: Learning Force-Adaptive Humanoid Loco-Manipulation}, 
      author={Yuanhang Zhang and Yifu Yuan and Prajwal Gurunath and Tairan He and Shayegan Omidshafiei and Ali-akbar Agha-mohammadi and Marcell Vazquez-Chanlatte and Liam Pedersen and Guanya Shi},
      year={2025},
      eprint={2505.06776},
      archivePrefix={arXiv},
      primaryClass={cs.RO},
      url={https://arxiv.org/abs/2505.06776}, 
}

@INPROCEEDINGS{2024ICRA-Sim-to-Real,
  author={Dao, Jeremy and Duan, Helei and Fern, Alan},
  booktitle={2024 IEEE International Conference on Robotics and Automation (ICRA)}, 
  title={Sim-to-Real Learning for Humanoid Box Loco-Manipulation}, 
  year={2024},
  volume={},
  number={},
  pages={16930-16936},
  doi={10.1109/ICRA57147.2024.10610977}}

@INPROCEEDINGS{2024ICRA-Learning_Force_Contro,
  author={Portela, Tifanny and Margolis, Gabriel B. and Ji, Yandong and Agrawal, Pulkit},
  booktitle={2024 IEEE International Conference on Robotics and Automation (ICRA)}, 
  title={Learning Force Control for Legged Manipulation}, 
  year={2024},
  volume={},
  number={},
  pages={15366-15372},
  doi={10.1109/ICRA57147.2024.10611066}}

@inproceedings{2024CoRL-Human_Plus,
title={HumanPlus: Humanoid Shadowing and Imitation from Humans},
author={Zipeng Fu and Qingqing Zhao and Qi Wu and Gordon Wetzstein and Chelsea Finn},
booktitle={8th Annual Conference on Robot Learning},
year={2024},
url={https://openreview.net/forum?id=WnSl42M9Z4}
}

@INPROCEEDINGS{2025ICRA-Mobile-TeleVision,
  author={Lu, Chenhao and Cheng, Xuxin and Li, Jialong and Yang, Shiqi and Ji, Mazeyu and Yuan, Chengjing and Yang, Ge and Yi, Sha and Wang, Xiaolong},
  booktitle={2025 IEEE International Conference on Robotics and Automation (ICRA)}, 
  title={Mobile-TeleVision: Predictive Motion Priors for Humanoid Whole-Body Control}, 
  year={2025},
  volume={},
  number={},
  pages={5364-5371},
  doi={10.1109/ICRA55743.2025.11128652}}

@INPROCEEDINGS{2023ICRA-Hierarchical_Adaptiv,
  author={Sombolestan, Mohsen and Nguyen, Quan},
  booktitle={2023 IEEE International Conference on Robotics and Automation (ICRA)}, 
  title={Hierarchical Adaptive Loco-manipulation Control for Quadruped Robots}, 
  year={2023},
  volume={},
  number={},
  pages={12156-12162},
  doi={10.1109/ICRA48891.2023.10160523}}

@INPROCEEDINGS{2024ICRA-Hierarchical_Optimization-based,
  author={Rigo, Alberto and Hu, Muqun and Gupta, Satyandra K. and Nguyen, Quan},
  booktitle={2024 IEEE International Conference on Robotics and Automation (ICRA)}, 
  title={Hierarchical Optimization-based Control for Whole-body Loco-manipulation of Heavy Objects}, 
  year={2024},
  volume={},
  number={},
  pages={15322-15328},
  doi={10.1109/ICRA57147.2024.10611656}}

@INPROCEEDINGS{2025ICRA-Full-Order,
  author={Xue, Haoru and Pan, Chaoyi and Yi, Zeji and Qu, Guannan and Shi, Guanya},
  booktitle={2025 IEEE International Conference on Robotics and Automation (ICRA)}, 
  title={Full-Order Sampling-Based MPC for Torque-Level Locomotion Control via Diffusion-Style Annealing}, 
  year={2025},
  volume={},
  number={},
  pages={4974-4981},
  doi={10.1109/ICRA55743.2025.11127320}}

@article{HECTOR,
  title={Dynamic Loco-manipulation on HECTOR: Humanoid for Enhanced ConTrol and Open-source Research},
  author={Li, Junheng and Ma, Junchao and Kolt, Omar and Shah, Manas and Nguyen, Quan},
  journal={arXiv preprint arXiv:2312.11868},
  year={2023}
}

@misc{Kinodynamics,
      title={Kinodynamics-based Pose Optimization for Humanoid Loco-manipulation}, 
      author={Junheng Li and Quan Nguyen},
      year={2023},
      eprint={2303.04985},
      archivePrefix={arXiv},
      primaryClass={cs.RO},
      url={https://arxiv.org/abs/2303.04985}, 
}

@INPROCEEDINGS{2005ICRA,
  author={Harada, K. and Kajita, S. and Saito, H. and Morisawa, M. and Kanehiro, F. and Fujiwara, K. and Kaneko, K. and Hirukawa, H.},
  booktitle={Proceedings of the 2005 IEEE International Conference on Robotics and Automation}, 
  title={A Humanoid Robot Carrying a Heavy Object}, 
  year={2005},
  volume={},
  number={},
  pages={1712-1717},
  doi={10.1109/ROBOT.2005.1570360}}

@misc{FACET,
      title={FACET: Force-Adaptive Control via Impedance Reference Tracking for Legged Robots}, 
      author={Botian Xu and Haoyang Weng and Qingzhou Lu and Yang Gao and Huazhe Xu},
      year={2025},
      eprint={2505.06883},
      archivePrefix={arXiv},
      primaryClass={cs.RO},
      url={https://arxiv.org/abs/2505.06883}, 
}

@article{2024CoRL-OmniH2O,
  publtype={informal},
  author={Tairan He and Zhengyi Luo and Xialin He and Wenli Xiao and Chong Zhang and Weinan Zhang and Kris Kitani and Changliu Liu and Guanya Shi},
  title={OmniH2O: Universal and Dexterous Human-to-Humanoid Whole-Body Teleoperation and Learning},
  year={2024},
  cdate={1704067200000},
  journal={CoRR},
  volume={abs/2406.08858},
  url={https://doi.org/10.48550/arXiv.2406.08858}
}

@INPROCEEDINGS{2025ICRA-HOVER,
  author={He, Tairan and Xiao, Wenli and Lin, Toru and Luo, Zhengyi and Xu, Zhenjia and Jiang, Zhenyu and Kautz, Jan and Liu, Changliu and Shi, Guanya and Wang, Xiaolong and Fan, Linxi Jim and Zhu, Yuke},
  booktitle={2025 IEEE International Conference on Robotics and Automation (ICRA)}, 
  title={HOVER: Versatile Neural Whole-Body Controller for Humanoid Robots}, 
  year={2025},
  volume={},
  number={},
  pages={9989-9996},
  doi={10.1109/ICRA55743.2025.11128549}}

@misc{Clone,
      title={CLONE: Closed-Loop Whole-Body Humanoid Teleoperation for Long-Horizon Tasks}, 
      author={Yixuan Li and Yutang Lin and Jieming Cui and Tengyu Liu and Wei Liang and Yixin Zhu and Siyuan Huang},
      year={2025},
      eprint={2506.08931},
      archivePrefix={arXiv},
      primaryClass={cs.RO},
      url={https://arxiv.org/abs/2506.08931}, 
}

@misc{2025RSS-HOMIE,
      title={HOMIE: Humanoid Loco-Manipulation with Isomorphic Exoskeleton Cockpit}, 
      author={Qingwei Ben and Feiyu Jia and Jia Zeng and Junting Dong and Dahua Lin and Jiangmiao Pang},
      year={2025},
      eprint={2502.13013},
      archivePrefix={arXiv},
      primaryClass={cs.RO},
      url={https://arxiv.org/abs/2502.13013}, 
}

@misc{PPO,
      title={Proximal Policy Optimization Algorithms}, 
      author={John Schulman and Filip Wolski and Prafulla Dhariwal and Alec Radford and Oleg Klimov},
      year={2017},
      eprint={1707.06347},
      archivePrefix={arXiv},
      primaryClass={cs.LG},
      url={https://arxiv.org/abs/1707.06347}, 
}

@InProceedings{AMASS,
author = {Mahmood, Naureen and Ghorbani, Nima and Troje, Nikolaus F. and Pons-Moll, Gerard and Black, Michael J.},
title = {AMASS: Archive of Motion Capture As Surface Shapes},
booktitle = {Proceedings of the IEEE/CVF International Conference on Computer Vision (ICCV)},
month = {October},
year = {2019}
}

@ARTICLE{HAFO,
  author={Dong, Chenhui and Xu, Haozhe and Feng, Wenhao and Wang, Zhipeng and Zhou, Yanmin and Zhao, Yifei and He, Bin},
  journal={Control Engineering Practice},
  title={HAFO: A Force-Adaptive Control Framework for Humanoid Robots in Intense Interaction Environments},
  year={2026},
  volume={173},
  pages={106978},
  doi={10.1016/j.conengprac.2026.106978}
}

@misc{FAME,
      title={FAME: Force-Adaptive RL for Expanding the Manipulation Envelope of a Full-Scale Humanoid},
      author={Niraj Pudasaini and Yutong Zhang and Jensen Lavering and Alessandro Roncone and Nikolaus Correll},
      year={2026},
      eprint={2603.08961},
      archivePrefix={arXiv},
      primaryClass={cs.RO},
      url={https://arxiv.org/abs/2603.08961},
}

@misc{HEFT,
      title={HEFT: Heavy-Payload Full-size Humanoid Teleoperation with Privileged Motion Guidance and Windowed Payload Curriculum},
      author={Chenxin Liu and Qingzhou Lu and Guangxiao Yang and Xuanyang Shi and Chenghan Yang and Yanjiang Guo and Jianyu Chen},
      year={2026},
      eprint={2607.02332},
      archivePrefix={arXiv},
      primaryClass={cs.RO},
      url={https://arxiv.org/abs/2607.02332},
}

@article{CWI,
  author  = {Ge, Wenqi and Guo, Junde and Fu, Zhen and Yang, Shunpeng and Chen, Jiayu and Chen, Hua},
  title   = {CWI: Composite Humanoid Whole-Body Imitation System for Loco-Manipulation},
  journal = {IEEE Robotics and Automation Letters},
  year    = {2026},
  doi     = {10.1109/LRA.2026.3710366}
}

@misc{HAF,
  title        = {HAF: Adapting Generalist VLAs to Humanoid Whole-Body Loco-manipulation via Hierarchical Action Flow and Spectral Latent RL},
  author       = {Langzhe Gu and Chengkai Hou and Meng Li and Xinhua Wang and Jiaming Liu and Xinyuan Lv and Bowei Zhang and Shuanghao Bai and Guangrun Li and Jingyang He and Gaole Dai and Ziluo Ding and Zhiyuan Xu and Kuan Cheng and Jian Tang and Zhengping Che and Shanghang Zhang},
  year         = {2026},
  eprint       = {2608.16837},
  archivePrefix= {arXiv},
  primaryClass = {cs.RO}
}

@article{SONIC,
author = {Zhengyi Luo  and Ye Yuan  and Tingwu Wang  and Chenran Li  and Fernando Casta{\~n}eda  and Sirui Chen  and Zi-Ang Cao  and Jiefeng Li  and David Minor  and Qingwei Ben  and Jinhyung Park  and David Sami  and Zi Wang  and Xingye Da  and Runyu Ding  and Cyrus Hogg  and Lina Song  and Edy Lim  and Eugene Jeong  and Tairan He  and Haoru Xue  and Wenli Xiao  and Simon Yuen  and Jan Kautz  and Yan Chang  and Umar Iqbal  and Linxi "Jim" Fan  and Yuke Zhu },
title = {SONIC: Supersizing motion tracking for natural humanoid whole-body control},
journal = {Science Robotics},
volume = {11},
number = {117},
pages = {eaed4592},
year = {2026},
doi = {10.1126/scirobotics.aed4592},
URL = {https://www.science.org/doi/abs/10.1126/scirobotics.aed4592},
eprint = {https://www.science.org/doi/pdf/10.1126/scirobotics.aed4592}
}

@misc{Caron2016ZMPSupportAreas,
   author = {St{\'e}phane Caron and Quang-Cuong Pham and Yoshihiko Nakamura},
   title = {{ZMP} Support Areas for Multi-Contact Mobility under
            Frictional Constraints},
   year = {2016},
   eprint = {1510.03232},
   archivePrefix = {arXiv},
   primaryClass = {cs.RO},
   url = {https://arxiv.org/abs/1510.03232}
 }

@article{GMR,
  title={Retargeting Matters: General Motion Retargeting for Humanoid Motion Tracking},
  author= {Joao Pedro Araujo and Yanjie Ze and Pei Xu and Jiajun Wu and C. Karen Liu},
  year= {2025},
  journal= {arXiv preprint arXiv:2510.02252}
}

\end{document}